\documentclass[twoside]{article}
\usepackage[accepted]{aistats2018}
\usepackage{graphicx}
\usepackage{amsmath}
\usepackage{subfigure}
\usepackage{amssymb}
\usepackage{natbib}
\usepackage{mlmacros}
\usepackage{float}
\usepackage{multirow}
\newcommand{\eq}[1]{\begin{align*}#1\end{align*}}
\newcommand\numberthis{\addtocounter{equation}{1}\tag{\theequation}}
\variables[obs,latent]{x,z}
\probdists{p,q}

\newcommand{\elbo}{\mathcal{L}_{\text{ELBO}}}

\newcommand{\dist}[2]{\delta(#1, #2)}

\newcommand{\z}{\mathbf{z}}
\newcommand{\hatz}{\hat{\mathbf{z}}}
\newcommand{\x}{\mathbf{x}}
\newcommand{\J}{\mathbf{J}}
\newcommand{\G}{\mathbf{G}}
\newcommand{\U}{\mathbf{U}}
\newcommand{\s}{\mathbf{S}}
\newcommand{\V}{\mathbf{V}}
\newcommand{\h}{\mathbf{h}}
\newcommand{\w}{\mathbf{w}}
\newcommand{\cx}{f}
\newcommand{\cz}{\gamma}
\newcommand{\nn}{\omega}

\usepackage{todonotes}

\begin{document}

\graphicspath{{figures/}}

%

%

\twocolumn[

\aistatstitle{Metrics for Deep Generative Models}

\aistatsauthor{ Nutan Chen$^\ast$  \And Alexej Klushyn$^\ast$ \And  Richard Kurle$^\ast$  }
\aistatsauthor{Xueyan Jiang  \And  Justin Bayer  \And Patrick van der Smagt}
\aistatsaddress{ \texttt\{{first name dot last name\}@volkswagen.de} \\ AI Research, Data:Lab, Volkswagen Group, Munich, Germany 
 } 
]

\begin{abstract}
Neural samplers such as variational autoencoders (VAEs) or generative adversarial networks (GANs) approximate distributions by transforming samples from a simple random source---the latent space---to samples from  a more complex distribution represented by a dataset. 
While the manifold hypothesis implies that a dataset contains large regions of low density, the training criterions of VAEs and GANs will make the latent space densely covered. 
Consequently points that are separated by low-density regions in observation space will be pushed together in latent space, making stationary distances poor proxies for similarity.
We transfer ideas from Riemannian geometry to this setting, letting the distance between two points be the shortest path on a Riemannian manifold induced by the transformation. The method yields a principled distance measure, provides a tool for visual inspection of deep generative models, and an alternative to linear interpolation in latent space. In addition, it can be applied for robot movement generalization using previously learned skills.
The method is evaluated on a synthetic dataset with known ground truth; on a simulated robot arm dataset; on human motion capture data; and on a generative model of handwritten digits.

\end{abstract}

\section{Introduction}



Matrices of pairwise distances serve as the input for a wide range of classic machine learning algorithms, such as k-nearest neighbour, multidimensional scaling, or stationary kernels. 
In the case of high-dimensional spaces, obtaining a meaningful distance is challenging for two reasons. 
First, choosing a metric, e.g.\ an instance of Minkowski distance, comes with certain assumptions on the data---e.g., distances are invariant to rotation under the L2 norm. Second, these distances become increasingly meaningless for higher dimensions  \citep{aggarwal2001surprising}.
Numerous researchers have proposed to learn distances from data, sometimes referred to as \emph{metric learning} \citep{xing2003distance, weinberger2006distance, davis2007information, kulis2013metric}. 
For data distributed according to the multivariate normal, the Mahalanobis distance is a popular choice, making the distance measure effectively invariant to translation and scaling. 
The idea of a linear transformation of the data has been extended to also reflect supervised side information such as class labels \citep{WeinbergerS09,GoldbergerRHS04}. 
Further work has pushed this to use non-linear \citep{SalakhutdinovH07} and recurrent \citep{BayerOS12} transformations.

Probabilistic modelling of high-dimensional data has progressed enormously. Two distinct ``neural sampling'' approaches are those of generative adversarial networks (GANs) \citep{GoodfellowPMXWOCB14} and variational autoencoders (VAEs) \citep{KingmaW13,RezendeMW14}. 


This work aims to bring a set of techniques to neural sampling that makes them powerful tools for metric learning. Pioneering work on interpolation and generation between two given points on a Riemannian manifold includes \citep{noakes1989cubic} and \citep{crouch1995dynamic}.
In addition, Principal Geodesic Analysis (PGA) \citep{fletcher2004principal} describing the variability of data on a manifold uses geodesics in Principal component analysis. 
Recent work of \cite{tosi2014metrics} proposed to perceive the latent space of Gaussian process latent variable models (GP-LVMs) as a Riemannian manifold, where the distance between two data points is given as the shortest path along the data manifold. 

We transfer the idea of \citep{tosi2014metrics} to neural samplers. We show how to represent such shortest paths in a parameterized fashion, where the calculation of the distance between two data points is effectively a minimization of the length of a curve. The method is evaluated on a range of high-dimensional datasets. 
Further, we provide evidence of the manifold hypothesis \citep{RifaiDVBM11}. We would also like to mention \cite{arvanitidis2017latent} and \cite{shao2017riemannian} who independently worked on this topic at the same time as us.

In robotic domains, our approach can be applied to path planning based on the learned skills. The demonstrations from experts enable robots to generate particular motions. Simultaneously, robots require natural motion exploration \citep{havoutis2013motion}. In our method, a Riemannian metric is used to achieve such optimal motion paths along the data manifolds.



\section{Using Riemannian geometry in generative latent variable models}
\label{sec:motivation}

Latent variable models are commonly defined as 
\eq{
    \p{\bobs} = \int \p{\bobs}{\blatent}\,\p{\blatent} \,d\blatent , \numberthis \label{eq:nlvm}
}
where latent variables $\blatent \in \mathbb{R}^{N_z}$ are used to explain the data $\bobs \in \mathbb{R}^{N_x}$.

Assume we want to obtain a distance measure $\dist{\bobs}{\tilde{\bobs}}$ from the learned manifolds, which adequately reflects the ``similarity'' between data points $\bobs$ and $\tilde{\bobs}$.
If we can infer the corresponding latent variables $\blatent$ and $\tilde{\blatent}$, an obvious choice is the Euclidean distance in latent space, $||\blatent - \tilde{\blatent}||_2$.
This has the implicit assumption that moving a certain distance in latent spaces moves us proportionally far in observation space, $||\blatent - \tilde{\blatent}||_2 \propto  ||\bobs - \tilde{\bobs}||_2 $.
But this is a fallacy: for latent variables to adequately model the data, stark discontinuities in the likelihood $\p{\bobs}{\blatent}$ are virtually always present.
To see this, we note that the prior can be expressed as the posterior aggregated over the data:
\eq{
    \p{\blatent} &= \int \p{\blatent}{\bobs}\,\p{\bobs}\, d\bobs=\mathbb{E}_{\bobs \sim \p{\mathbf{X}}}[\p{\blatent}{\bobs}].
}

A direct consequence of the discontinuities is that there are no regions of low density in the latent space.
Hence, separated manifolds in the observation space (\eg the set of points from different classes) may be placed directly next to each other in latent space---a property that can only be compensated through rapid changes in the likelihood $\p{\bobs}{\blatent}$ at the respective ``borders".

For the estimation of nonlinear latent variable models we use importance-weighted autoencoders (IWAE) \citep{BurdaGS15} (see Section~\ref{sec:methods-iwae}). 
Treating the latent space as a Riemannian manifold (see Section~\ref{sec:methods:riemannian-geometry}) provides tools to define distances between data points by taking into account changes in the likelihood.

\subsection{Importance-weighted autoencoder}
\label{sec:methods-iwae}
Inference and learning in models of the form given by Eq.~(\ref{eq:nlvm})---based on the maximum-likelihood principle---are intractable because of the marginalization over the latent variables.
Typically, approximations are used which are either based on sampling or on variational inference.
In the latter case, the intractable posterior is approximated by a distribution $\q{\blatent^{(i)}}$.
The problem of inference is then substituted by one of optimization, namely the maximization of the evidence lower bound (ELBO).
Let $\mathbf{X} = \{ \mathbf{x}^{(1)},\dots,\mathbf{x}^{(N)} \}$ be observable data and $\mathbf{z}^{(i)}$ the corresponding latent variables.
Further, let $p_\theta(\bobs|\blatent)$ be a likelihood function parameterized by $\theta$.
Then
\begin{equation}
\begin{aligned}
& \ln p_{\theta}(\mathbf{X})
= \sum_{i=1}^{N} \ln p_{\theta}(\mathbf{x}^{(i)}) \\
&  \geq \sum_{i=1}^{N} \mathbb{E}_{q(\mathbf{z}^{(i)})} \Big{[} \ln  \frac{p_{\theta}(\mathbf{x}^{(i)} | \mathbf{z}^{(i)}) p_{\theta}(\mathbf{z}^{(i)})}{q(\mathbf{z}^{(i)})} \Big{]} = \elbo .
\end{aligned}
\end{equation}

If we implement $\q{\blatent^{(i)}} = q_{\phi}(\blatent|\bobs^{(i)})$ with a neural network parameterized by $\phi$, we obtain the variational autoencoder of \cite{KingmaW13}, which jointly optimizes $\elbo$ with respect to $\theta$ and $\phi$.
Since the inference and generative models are tightly coupled, an inflexible variational posterior has a direct impact on the generative model, causing both models to underuse their capacity. 
In order to learn richer latent representations and achieve better generative performance, the importance-weighted autoencoder (IWAE) \citep{BurdaGS15, cremer2017reinterpreting} has been introduced. It treats $q_{\phi}(\mathbf{z} | \mathbf{x})$ as a proposal distribution and obtains a tighter lower bound using importance sampling:
\begin{equation}
\label{eq:K}
\begin{aligned}
& \ln p_{\theta}(\mathbf{X}) = \sum_{i=1}^{N} \ln p_{\theta}(\mathbf{x}^{(i)}) \\
& \geq \sum_{i=1}^{N} \mathbb{E}_{\mathbf{z}^{(i)}_{1},\dots , \mathbf{z}^{(i)}_{K} \sim q_{\phi}(\mathbf{z}^{(i)} | \mathbf{x}^{(i)})} \Big{[} 
\ln \frac{1}{K} \sum_{k=1}^{K} w^{(i)}_{k} \Big{]},
\end{aligned}
\end{equation}
where $w^{(i)}_{k}$ are the importance weights:
\begin{equation}
w^{(i)}_{k} = \frac{p_{\theta}(\mathbf{x}^{(i)} | \mathbf{z}^{(i)}_{k}) \,p_{\theta}(\mathbf{z}^{(i)}_{k})}{q_{\phi}(\mathbf{z}^{(i)}_{k} | \mathbf{x}^{(i)})}.
\end{equation}
The IWAE is the basis of our approach, since it can yield an accurate generative model.

\subsection{Riemannian geometry}
\label{sec:methods:riemannian-geometry}
A Riemannian manifold is a differentiable manifold $M$ with a metric tensor $\G$. It assigns to each point $\mathbf{\z}$ an inner product on the tangent space $T_{\mathbf{\z}} M$, where the inner product is defined as:
\begin{equation}
\langle\z', \z'\rangle_{\z} := \z'^{T}\, \G(\z)\, \z'
\end{equation}
with $\z' \in T_{\z}M$ and $\z \in M$.

Consider a curve $\cz:[0, 1]\rightarrow\mathbb{R}^{N_{z}}$ in the Riemannian manifold, transformed by a continuous function $\cx(\cz(t))$ to an $N_{x}$-dimensional observation space, where $\cz(t)\in\mathbb{R}^{N_{z}}$.
The length of the curve in the observation space is defined as

\begin{align}
\label{eq:length}
L(\cz)&:=\int_0^1 \left\| \frac{\partial \cx(\cz(t))}{\partial t}\right\| \mathrm{d}t  \nonumber \\
&= \int_0^1 \left\| \frac{\partial \cx(\cz(t))}{\partial \cz(t)} \frac{\partial \cz(t)}{\partial t} \right\| \mathrm{d}t \nonumber \\
&=\int_0^1 \left\| \mathbf{J} \frac{\partial \gamma(t)}{\partial t} \right\| \mathrm{d}t ,
\end{align}
where $\mathbf{J}$ is the Jacobian.
Eq.\ (\ref{eq:length}) can be expressed as
\begin{align} \label{eq:length_final}
L(\cz) = \int_0^1\sqrt{\big<\gamma'(t), \gamma'(t)\big>_{\gamma(t)}}\mathrm{d}t
\end{align}
with the metric tensor $\G=\mathbf{J}^{T}\mathbf{J}$.

\section{Approximating the geodesic}
\label{sec:methods}

In this work, we are primarily interested in length-minimizing curves between samples of generative models.
In Riemannian geometry, locally length-minimizing curves are referred to as geodesics.
We treat the latent space of generative models as a Riemannian manifold.
This allows us to parametrize the curve in the latent space, while distances are measured by taking into account distortions from the generative model.

We use a neural network $g_{\nn}:\mathbb{R} \rightarrow \mathbb{R}^{N_{z}}$ to approximate the curve $\cz$ in the latent space, where $\nn$ are the weights and biases.

The function $\cx$ from Eqs.\ \eqref{eq:length} corresponds to the mean of the generative model's probability distribution $h^\mathrm{gen}:\mathbb{R}^{N_{z}} \rightarrow \mathbb{R}^{N_{x}}$ and the components of the Jacobian are

 \begin{align}
 J_{i,j} = \frac{\partial x_i}{\partial z_j}.
 \end{align}
$x_i$ and $z_j$ denote the $i$-th and $j$-th element of the generated data points $\x$ and latent variables $\z$, respectively, with $\x \in \mathbb{R}^{N_{x}}$ and $\z \in \mathbb{R}^{N_{z}}$.

We approximate the integral of Eq.~(\ref{eq:length_final}) with $n$ equidistantly spaced sampling points of $t \in [0,1]$:
\begin{align}
L(g_{\nn}(t)) &\approx \frac{1}{n}\sum_{i=1}^n \sqrt{\langle g_{\nn}'(t_{i}), g_{\nn}'(t_{i})\rangle_{g_{\nn}(t_{i})}} \\
&= \frac{1}{n}\sum_{i=1}^n
\sqrt{g_{\nn}'(t_{i})^{T} \mathbf{J}^{T} \mathbf{J} g_{\nn}'(t_{i})}.
\label{eq:length_approx}
\end{align}
The term inside the summation can be interpreted as the rate of change at point $g_{\nn}(t_{i})$, induced by the generative model, and we will refer to it as velocity:
\begin{align}
\phi(t) = \sqrt{\langle g_{\nn}'(t_{i}), g_{\nn}'(t_{i})\rangle_{g_{\nn}(t_{i})}} .
\label{eq:velocity}
\end{align}
An approximation of the geodesic between two points in the latent space is obtained by minimizing the length in Eq.~(\ref{eq:length_approx}), where the weights and biases $\nn$ of the neural network $g_{\nn}(t)$ are subject to optimization.

With the start and end points of the curve in the latent space given as $\z_0$ and $\z_1$, we consider the following constrained optimization problem:
\begin{align}
\label{eq:constrained_optimization}
\min_{\nn} \ & L(g_{\nn}(t)) \nonumber \\
s.t. \ & g_{\nn}(0) = \z_{0}, \ g_{\nn}(1) = \z_{1}.
\end{align}

\subsection{Dealing with boundary constraints}

To satisfy the boundary constraints in Eq.~\eqref{eq:constrained_optimization}, we shift and rescale the predicted line to get
\begin{align}
\z (t)=\mathbf{A}\hatz (t)-\mathbf{B},
\end{align}
where
\begin{align}
\mathbf{A} &= \frac{\z_0-\z_1}{\hatz(0)-\hatz(1)} \nonumber \\  
\mathbf{B} &= \frac{\z_0\hatz(1)-\z_1\hatz(0)}{\hatz(0)-\hatz(1)}.
\end{align}
$\hat{\z}$ and $\z$ are the outputs of the neural network $g_{\nn}$ before cq.\ after normalization. The advantage of this proceed is that the optimization problem in Eq.~\eqref{eq:constrained_optimization} simplifies to find $\min_{\nn}L(g_{\nn}(t))$.

\subsection{Smoothing the metric tensor}
To ensure the geodesic is following the data manifold---which entails that the manifold distance is smaller than the Euclidean distance---a penalization term is added to smooth the metric tensor $\G$.
It leads to the following loss function:
\begin{align}
\mathcal{L}= L  + \lambda_{s} \|  \G \|_{2},
\end{align}
where $ \lambda_{s} > 0$ acts as a regularization coefficient. This optimization step is implemented as a post-processing of Eq.~(\ref{eq:constrained_optimization}) via singular-value decomposition (SVD)
\begin{align}
\G=\U \s \V^T,
\end{align}
where the columns of $\U$ are the eigenvectors of the covariance matrix $\G \G^{T}$, and the columns of V are the eigenvectors of $\G^{T} \G$. The diagonal entries in $\s$ contain singular values with scaling information about how a vector is stretched or shrunk when it is transformed from the column space of $\G$ to the row space of $\G$.  

Minimizing the term $ \lambda_{s} \|  \G \|_{2}$ is equivalent to a low-rank reconstruction for $\G$
\begin{align}
\hat{\G}=\U_r \mathrm{diag} \left\{ \frac{s_i^3}{s_i^2+\lambda_s} \right\}^{r}_{i=1} \V_r^T,
\end{align}
where $r$ is a pre-defined lower rank of $\G$. $r$ is predefined which is equal or slightly smaller than the full rank of the metric tensor. $\lambda_{s}$ rescales the singular values of $\G$ nonlinearly, which allows making the smaller singular values much smaller than the leading singular values. The smoothing therefore weakens the reconstructed off-diagonal values of $\hat{\G}$ which correspondingly reduces the manifold distance dramatically compared to the Euclidean distance. The smoothing effect is that a higher $\lambda_{s}$ augments the difference between the Euclidean interpolation and the path along the manifold---and will be demonstrated experimentally in Section~\ref{sec:pendulum}.

\section{Experiments}
\label{sec:experiments}

We evaluate our approach by conducting a series of experiments on three different datasets---an artificial pendulum dataset, the binarized MNIST digit dataset \citep{larochelle2011neural}, a simulated robot arm dataset and the human motion dataset\footnote{http://mocap.cs.cmu.edu/}.

Our goal is to enable smooth interpolations between the reconstructed images of an importance-weighted autoencoder and to differentiate between classes within the latent space.
To show that the paths of geodesics can differ from Euclidean interpolations, the following experiments mainly focus on comparing geodesics with the Euclidean interpolations as well as the reconstructed data generated from points along their paths.

\subsection{Training}

In all experiments, we chose a Gaussian prior $p(\mathbf{z}) = \mathcal{N}(\mathbf{0}, \mathbf{I})$. The inference model and the likelihood are represented by random variables of which the parameters are functions of the respective conditions.

For the inference model we consistently used a diagonal Gaussian, i.e.\  $q_{\phi}(\blatent | \bobs) = \mathcal{N}(\mu_{\phi}(\bobs), \text{diag}(\sigma_{\phi}^2(\bobs)))$. Depending on the experiments, the likelihood $p_{\theta}(\bobs|\blatent)$ either represents a Bernoulli variable $\mathcal{B}(r_{\theta}(\blatent))$ or a Gaussian $\mathcal{N}(\mu_{\theta}(\blatent), \sigma^2)$. $\sigma$ is a global variable and the parameters $r_{\theta}, \mu_{\theta}, \sigma_{\theta}^2, \mu_{\phi}, \sigma_{\phi}^2$ are functions of the latent variables represented by neural networks parameterized by $\theta$ and $\phi$ respectively.

The hyperparameters of $g_{\nn}$ are summarized in Table~\ref{table:geodesic}.
We used sigmoid, tanh and softplus activation functions in the generative model (see App.~\ref{appendix:gradients} and \ref{appendix:nonlinear_gradient}). See App.~\ref{appendix:training_procedure} for further details of the training procedure.

\subsection{Visualization}

There are several approaches to visualize the properties of the metric tensor, including Tissot's indicatrix. We use the magnification factor to visualize metric tensors during the evaluation, when we have two latent dimensions. The magnification factor \citep{bishop1997magnification} is defined as
\begin{align}
M\!F:=\sqrt{\det\G}.
\end{align}

\begin{figure*}[!ht]
	\centering
	\includegraphics[width=0.85\textwidth]{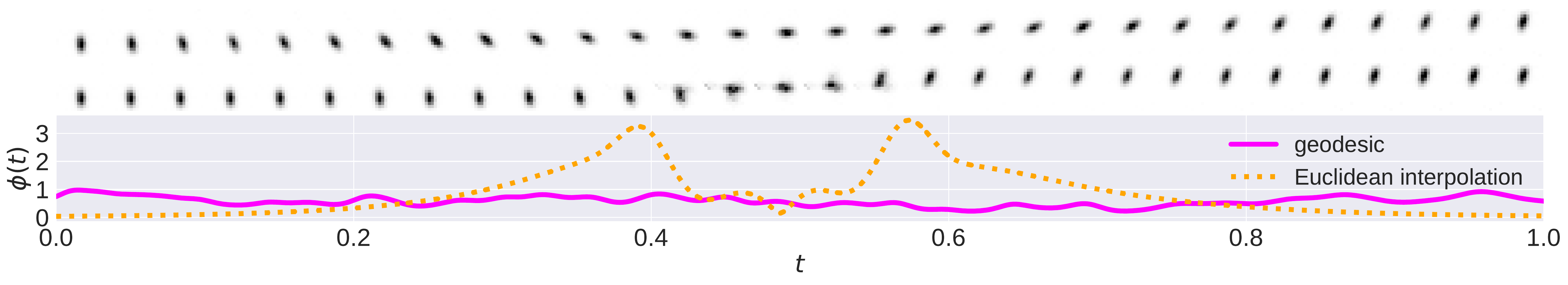}
	\caption{The reconstructions of the geodesic and the Euclidean interpolation between a single pair from the pendulum data. Top row: the mean of the reconstruction from the geodesic. Middle row: the mean of the reconstruction from the Euclidean interpolation. Bottom row: the velocity (Eq.~(\ref{eq:velocity})) of each sample. The distance of the Euclidean interpolation is 0.827, whereas the distance of the geodesic is 0.538.}
	\label{fig:pendulum_1example2}
\end{figure*}

To get an intuitive understanding of the magnification factor, it is helpful to consider the rule for changing variables $d\x=\sqrt{\det\J_\x}d\x'$. This rule shows the relation between infinitesimal volumes of different equidimensional Euclidean spaces. The same rule can be applied to express the relationship between infinitesimal volumes of a Euclidean space and a Riemannian manifold---with the difference of using the $M\!F$ instead of $\sqrt{\det\J_\x}$. Hence, the magnification factor visualizes the extent of change of the infinitesimal volume by mapping a point from the Riemannian manifold to the Euclidean space \citep{GemiciRM16}.

\subsection{Pendulum}
\label{sec:pendulum}

We created an artificial dataset of $16\times16$ pixel images of a pendulum with a joint angle as the only degree of freedom and augmented it by adding 0.05 per-pixel Gaussian noise. We generated 15,000 pendulum images, with joint angles uniformly distributed in the range [0, 360). The architecture of the IWAE can be found in Table~\ref{table:pendulum}.

\begin{figure}[!ht]
	\centering
	\includegraphics[width=0.4\textwidth]{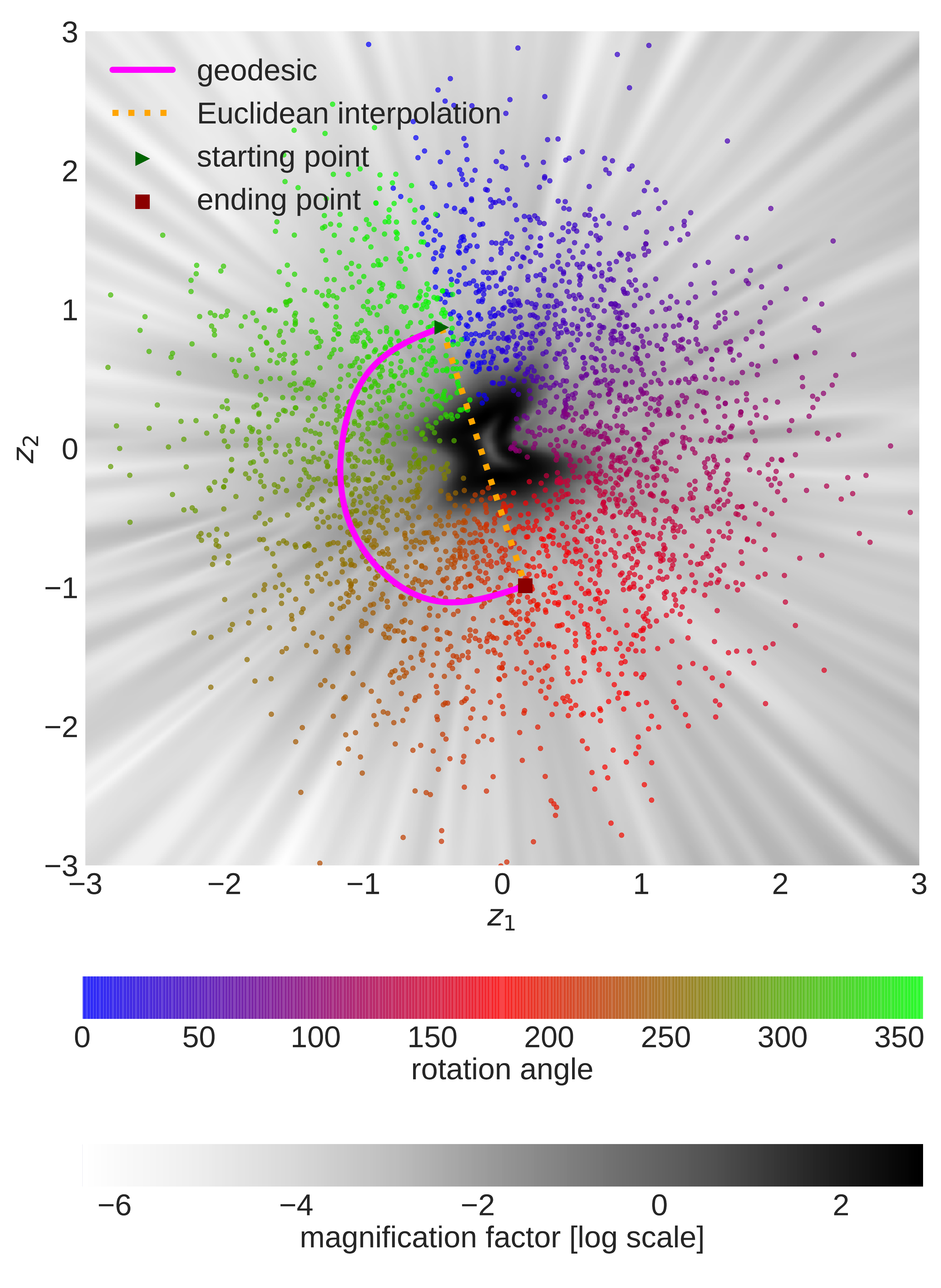}
	\caption{Geodesic and Euclidean interpolation in the latent space of the artificial pendulum dataset. The color encodes the pendulum rotation angles.}
	\label{fig:pendulum_1example}
\end{figure}

\begin{figure}[!ht]
	\centering
	\includegraphics[width=0.8\columnwidth]{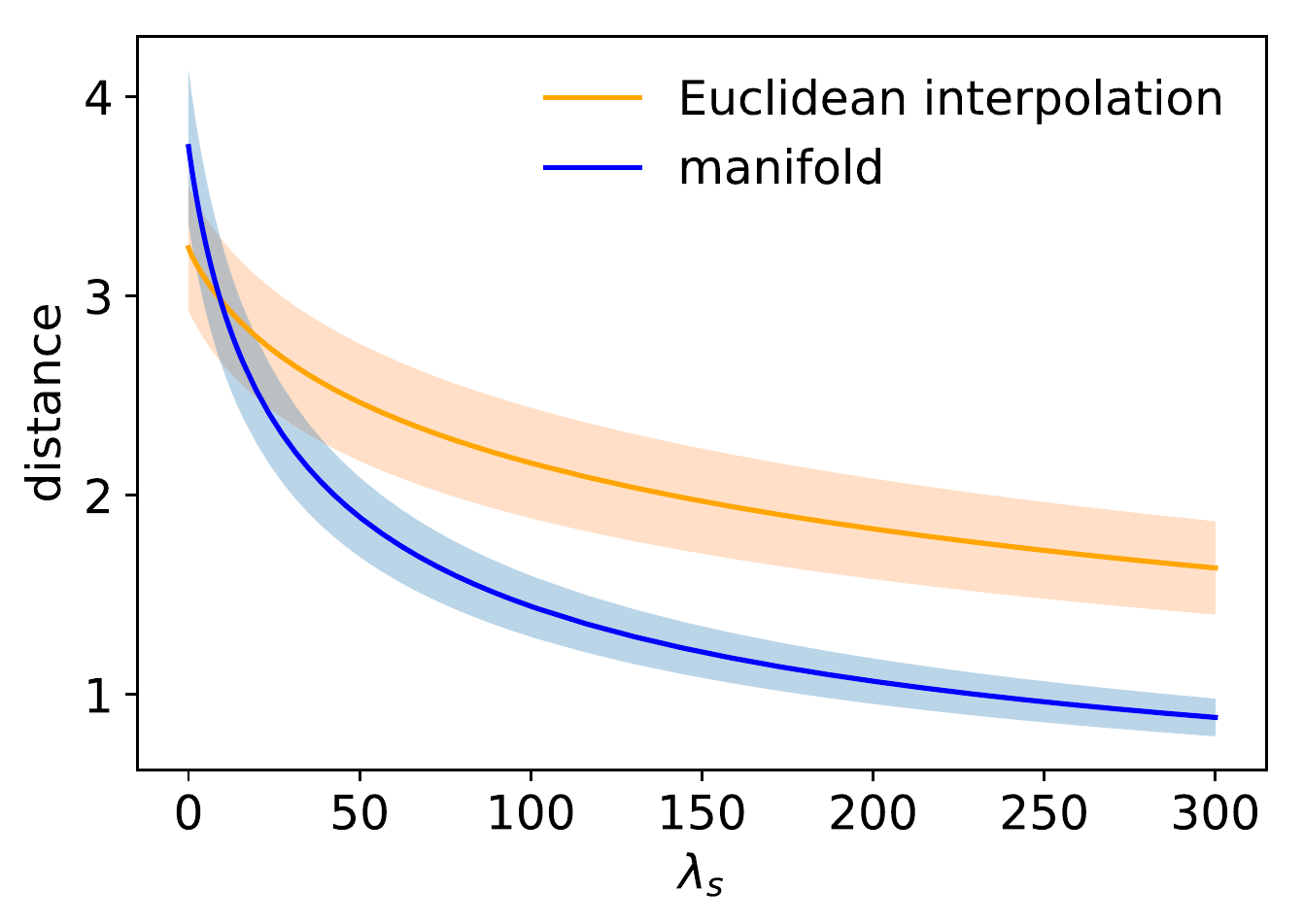}
	\caption{$\lambda_s$. The distances, along the Euclidean interpolation and the data manifold, between 100 pairs of randomly selected starting and end points. The shadow areas represent a 95\% confidence interval.}
	\label{fg:lambda_s}
\end{figure}

\begin{figure}[!ht]
	\centering
	\includegraphics[width=0.8\columnwidth]{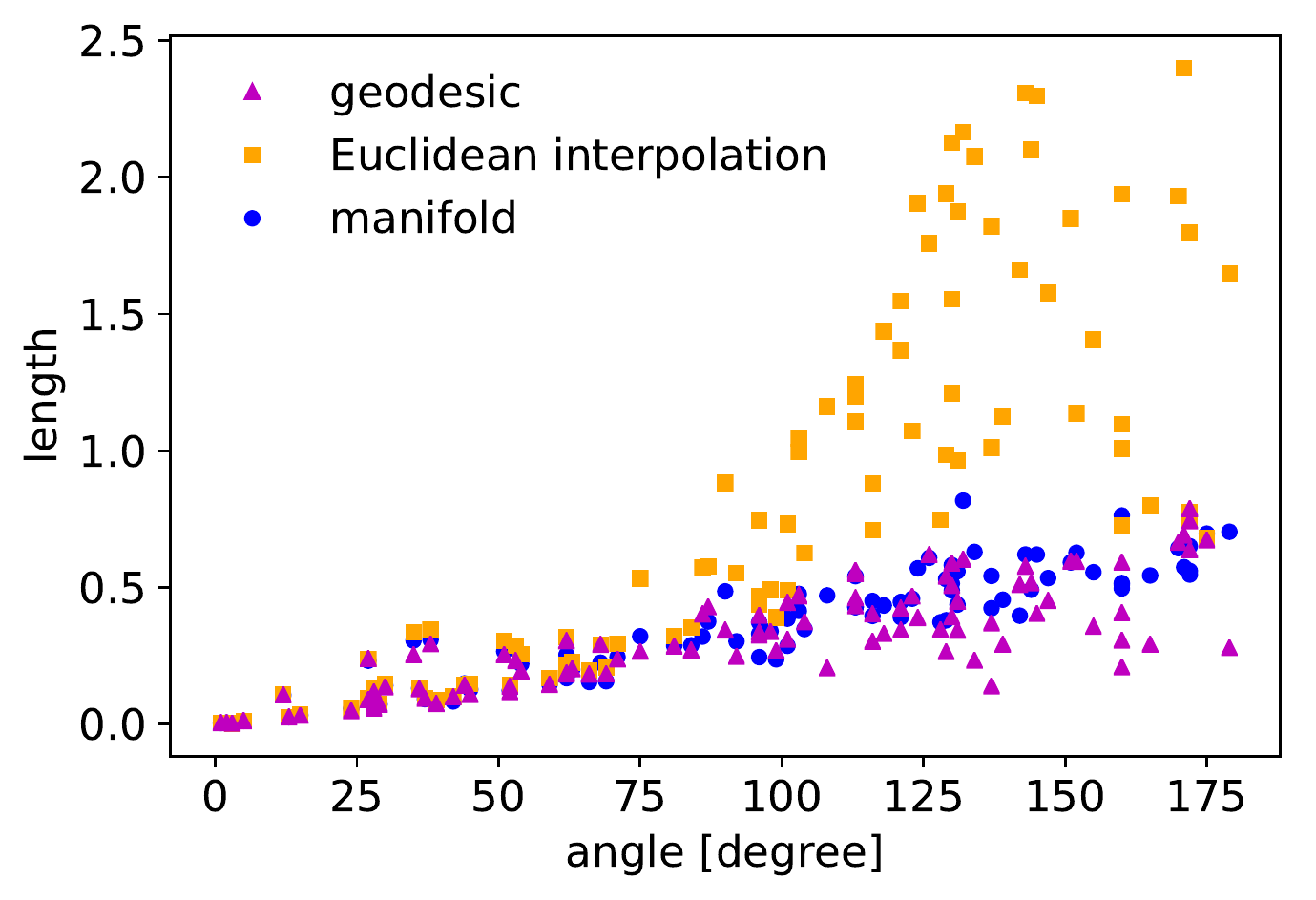}
	\caption{The horizontal axis is the angle between the starting and end points. The average of the length of the geodesics, the Euclidean interpolations, and the paths along the data manifold are 0.319, 0.820, and 0.353 respectively.}
	\label{fig:pendulum_100samples_lambda2000}
\end{figure}

\begin{figure*}[!ht]
	\centering
	\includegraphics[width=0.85\textwidth]{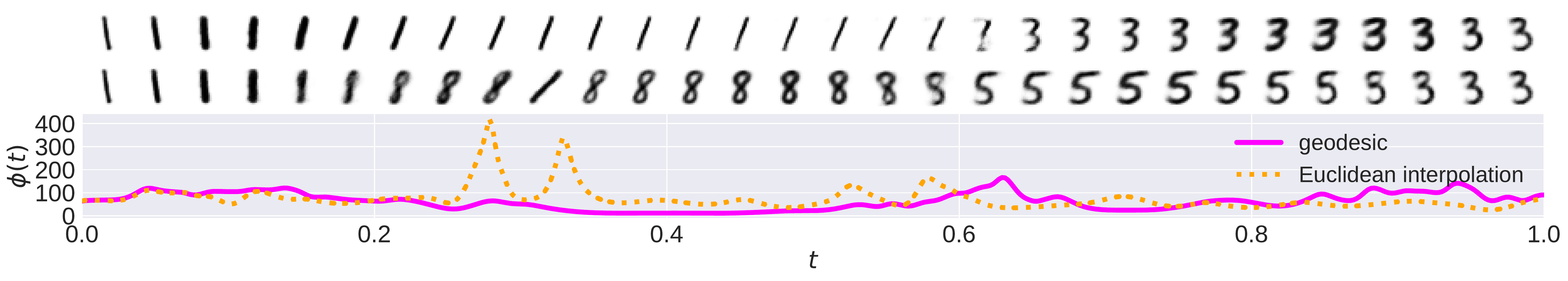}
	\caption{Reconstructions of the geodesic and the Euclidean interpolation based on the MNIST digit dataset. Top row: the mean of the reconstruction from the geodesic. Middle row: the mean of the reconstruction from the Euclidean interpolation. Bottom row: the velocity (Eq.~(\ref{eq:velocity})) of each sample. The distance of the Euclidean interpolation is 74.3, whereas the distance of the geodesic amounts to 62.9.}
	\label{fig:geo_euc_recs}
\end{figure*}

Fig.~\ref{fig:pendulum_1example} illustrates the trained two-dimensional latent space of the IWAE.
The grayscale in the background is proportional to the magnification factor, whereas the rotation angles of the pendulum are encoded by colors. The comparison of the geodesic (see Fig.~\ref{fig:pendulum_1example2}, top row) with the Euclidean interpolation (Fig.~\ref{fig:pendulum_1example2}, middle row) shows a much more uniform rotation of the pendulum for reconstructed images of points along the geodesic.

For this dataset, an SVD regularization with large values of $\lambda_s$ was necessary for the optimization, to yield a path along the data manifold. Fig.~\ref{fg:lambda_s} illustrates the influence of $\lambda_s$ on the distance metric. 
It is a property of this dataset, due to the generative distribution, that small values of $\lambda_s$ lead to shorter distances of the Euclidean interpolation than of paths along the data manifold. 
Fig.~\ref{fig:pendulum_100samples_lambda2000} shows the interpolations of 100 pairs of samples. We compared the geodesic with the Euclidean interpolation and an interpolation along the data manifold. The samples are randomly chosen with the condition to have a difference in the rotation angle of (0, 180]~degrees. The distances of the geodesics and the paths along the data manifold are linearly correlated to the angles between two points in the observation space and fit to each other.

\subsection{MNIST}

\begin{figure}[!ht]
	\centering
	\includegraphics[width=0.4\textwidth]{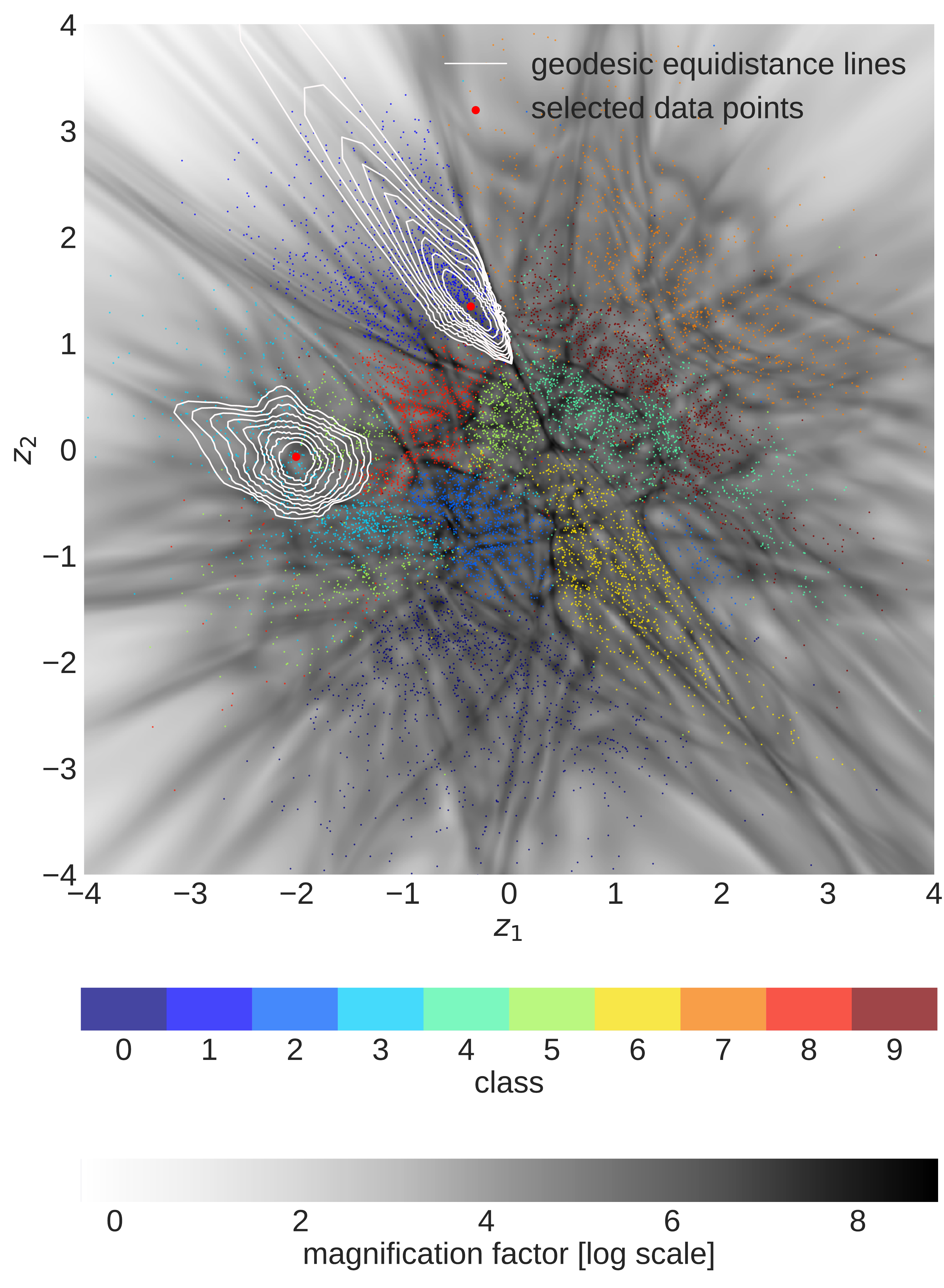}
	\caption{Equidistance lines around two selected data points in the latent space. Equidistance refers to the observation space and illustrates how regarding the latent space as a Riemannian manifold can help to separate classes.}
	\label{fig:equidistance_lines}
\end{figure}

To evaluate our model on a benchmark dataset, we used a fixed binarized version of the MNIST digit dataset defined by \cite{larochelle2011neural}. It consists of 50,000 training and 10,000 test images of handwritten digits (0 to 9) which are $28\times28$ pixels in size. The architecture of the IWAE is summarized in Table~\ref{table:mnist}.

\begin{figure}[!ht]
	\centering
	\includegraphics[width=0.4\textwidth]{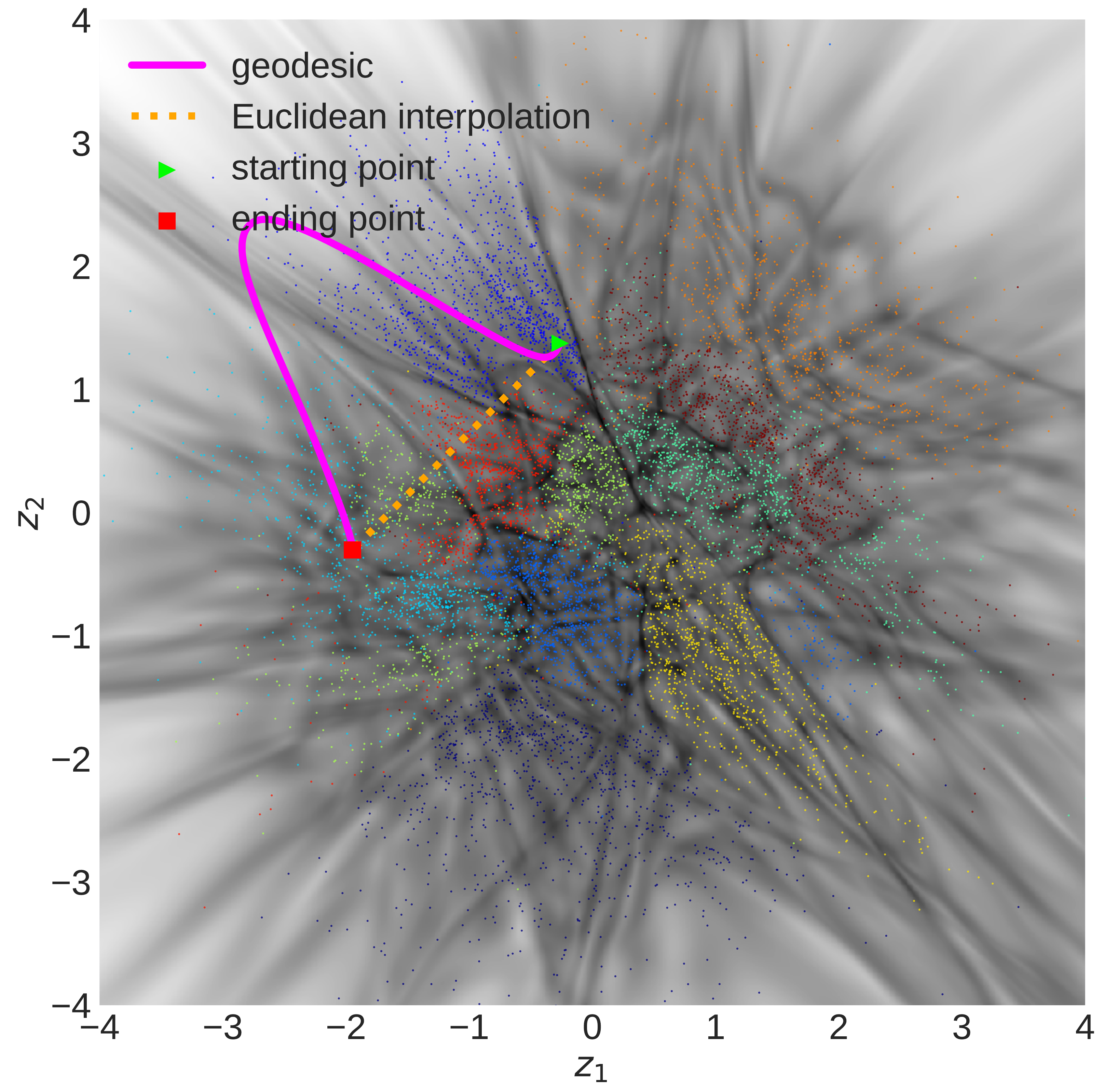}
	\caption{Geodesic and Euclidean interpolation in the latent space between two data points. The ten MNIST classes are encoded by colours, whereas the magnification factor is represented by the greyscale in the background. The $M\!F$ and the samples are  coloured the same as Fig.~\ref{fig:equidistance_lines}.}
	\label{fig:geo_euc_ls}
	\vspace{-0.1 in}
\end{figure}

Fig.~\ref{fig:equidistance_lines} shows the trained two-dimensional latent space of the IWAE. Distances between the selected data point and any point on the equidistance line are equal in the observation space. The courses of the equidistance lines demonstrate that treating the latent space as a Riemannian manifold enables to separate classes, since the geodesic between similar data points is shorter than between dissimilar ones. This is especially useful for state of the art methods that lead to very tight boundaries, like in this case---data points of different MNIST classes are almost not separable in the latent space by their Euclidean distance. Hence, the Euclidean distance cannot reflect the true similarity of two data points.

The difference between the geodesic and the Euclidean interpolation is shown in Fig.~\ref{fig:geo_euc_ls}. The Euclidean interpolation crosses four classes, the geodesic just two. Compared to the geodesic, the Euclidean interpolation leads to less smooth transitions in the reconstructions (see Fig.~\ref{fig:geo_euc_recs}, top and middle row). The transition between different classes is visualized by a higher velocity in this area (see Fig.~\ref{fig:geo_euc_recs}, bottom row). 

\subsection{Robot arm}

We simulated a movement of a KUKA robot that has six degrees of freedom (DOF). The end effector moved a circle with a 0.4 meter radius, generating a dataset with 6284 time steps. At each time step, the joint angles were obtained by inverse kinematics. The input data consisted of six-dimensional joint angles. Gaussian noise with a standard deviation of 0.03 was added to the data. The validation dataset also included a complete movement of a circle but only with 150 time steps. The architecture of the IWAE is shown in Table~\ref{table:robot}.

\begin{figure}[!ht]
	\centering
	\includegraphics[width=0.8\columnwidth]{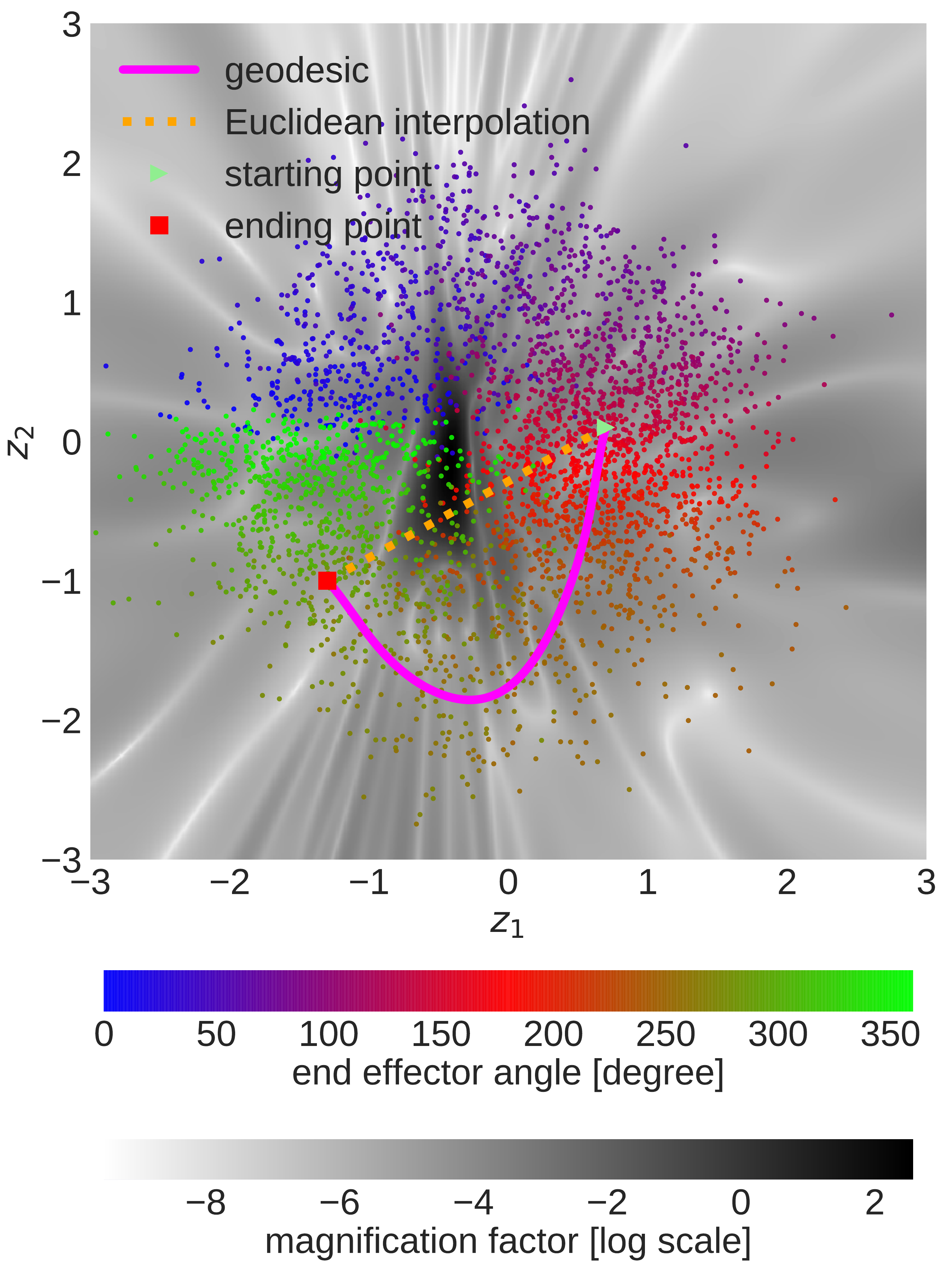}
	\caption{Geodesic and Euclidean interpolation in the latent space of the robot motions. }
	\label{fg:mf_robot}
\end{figure}

\begin{figure}[!ht]
	\centering
	\includegraphics[width=0.7\columnwidth]{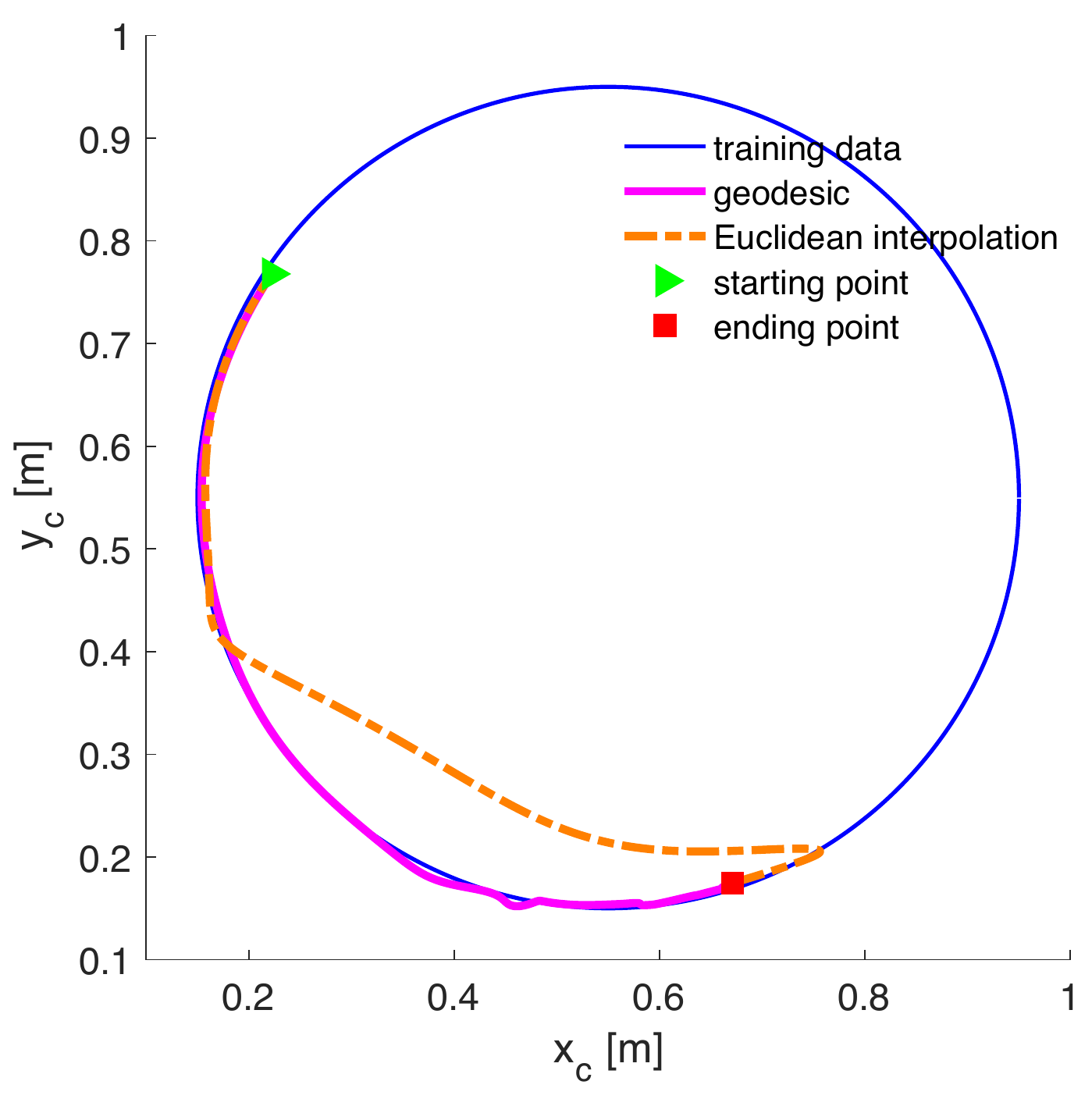}
	\caption{End effector trajectories in the Cartesian space. $x_c$ and $y_c$ represent the two axes of the end effector, while the third axis is not shown since the values on it are close to constant.}
	\label{fg:robot_end_effector}
\end{figure}

\begin{figure}[!ht]
	\centering
	\includegraphics[width=0.9\columnwidth]{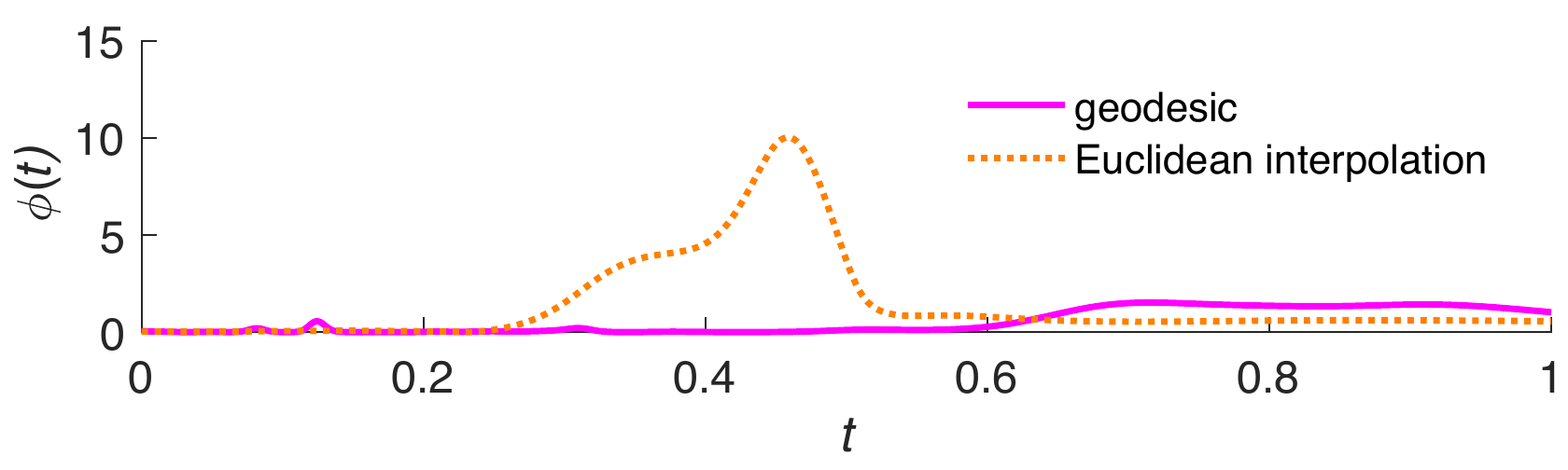}
	\caption{Velocity (Eq.~(\ref{eq:velocity})) of each robot sample. The distance of the Euclidean interpolation is 1.48, and the distance of the geodesic is 0.54.}
	\label{fg:robot_velocity}
\end{figure}

The geodesic interpolation outperforms the Euclidean interpolation for the robot arm movement, which is demonstrated in Fig.~\ref{fg:mf_robot}, \ref{fg:robot_end_effector} and \ref{fg:robot_velocity}.
For intuitive observations, the results are shown in a two-dimensional end effector Cartesian space using forward kinematics (see Fig.~\ref{fg:robot_end_effector}). 

To efficiently plan motions, in prior works \citep{berenson2009manipulation} constraints were created in the task space (e.g., constraint on the end-effector to move in a 2D instead of 3d). However, our method does not explicitly require these constraints.

The approach can be applied to movements with higher-dimensional joint angles like in case of
the full-body humanoids demonstrated in Section~\ref{sec:humanmotion}.

\subsection{Human motion}
\label{sec:humanmotion}

\begin{figure*}[!ht]
			\centering
			\includegraphics[width=0.85\textwidth]{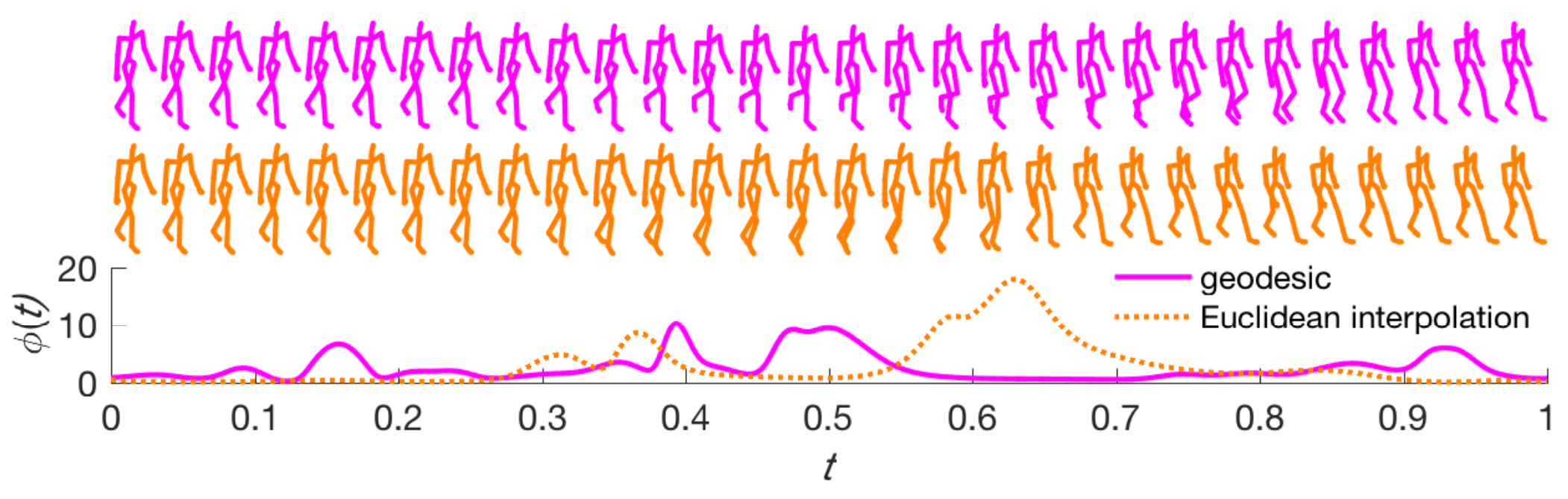}
			\caption{The reconstructions of the geodesic and Euclidean interpolation of the human motion. Top row: mean of the reconstruction from the geodesic. Middle row: mean of the reconstruction from the Euclidean interpolation. Bottom row: velocity (Eq.~(\ref{eq:velocity})) of each sample. The distance of the Euclidean interpolation is 2.89, and the distance of the geodesic is 2.57.}
			\label{fig:human_observation}
		\end{figure*}
		
\begin{figure}[!ht]
      \centering
       \includegraphics[width=0.4\textwidth]{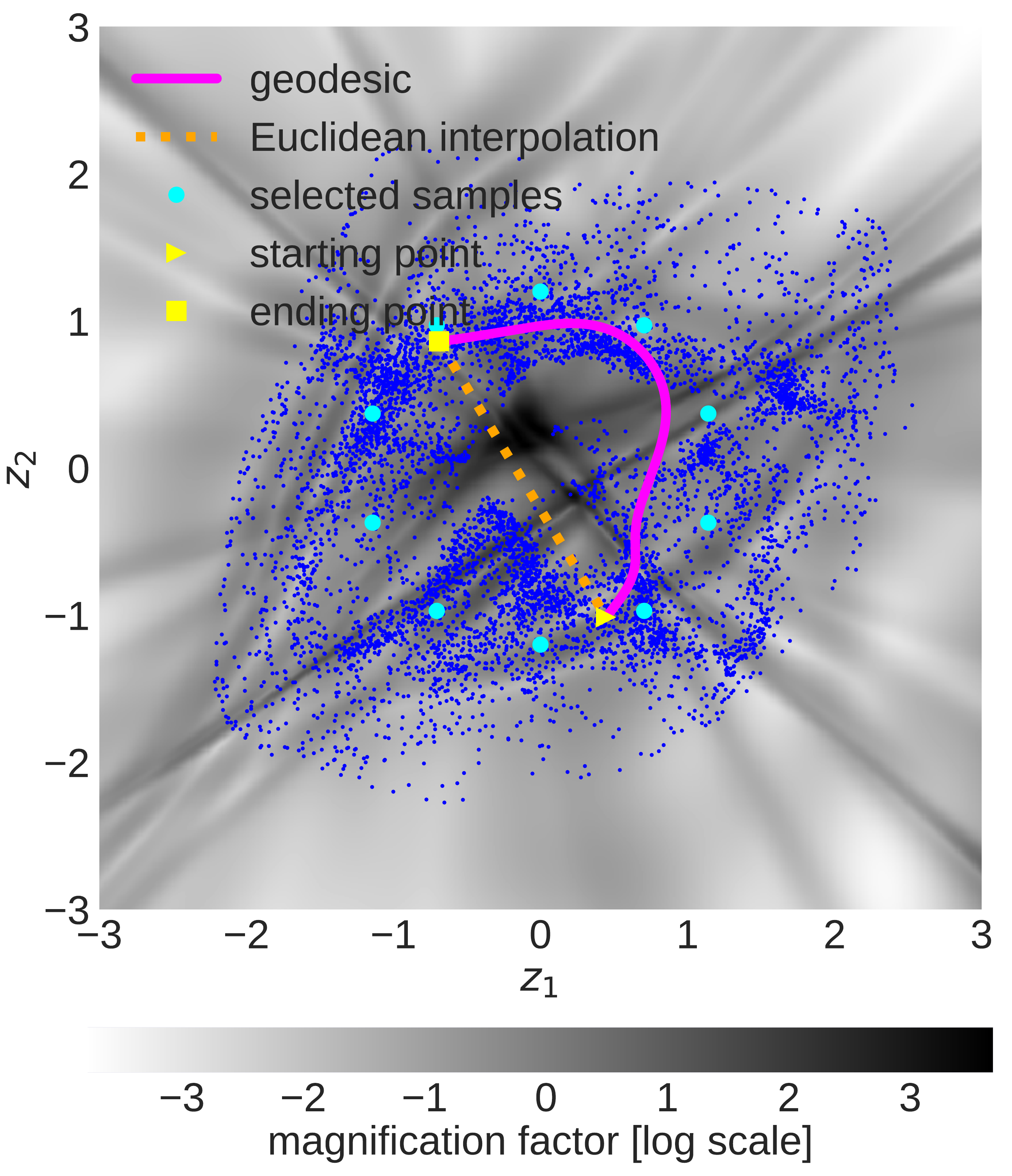}
	\caption{Geodesic and Euclidean interpolation in the latent space of the human motions. The blue dots are samples inferred from the training set.}
	\label{fig:human_latent}
	\vspace{-0.2 in}
\end{figure}

\begin{figure}[!ht]
      \centering
       \includegraphics[width=0.45\textwidth]{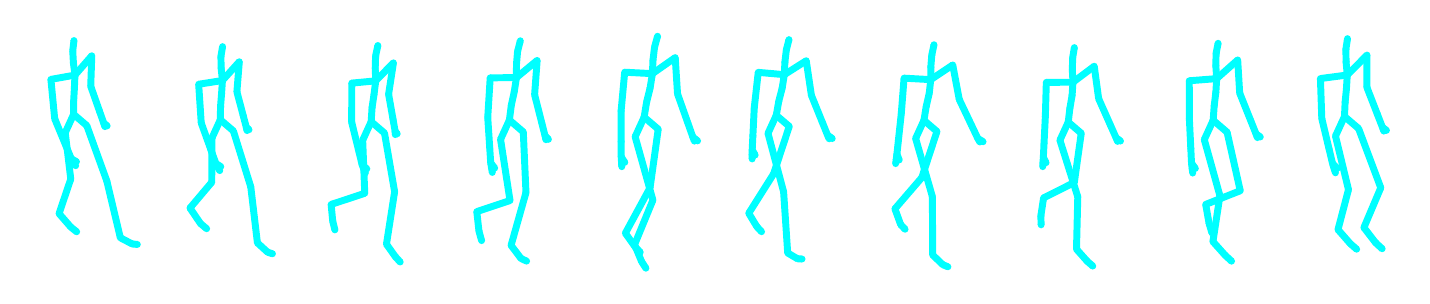}
	\caption{Selected samples of the human motion dataset. The subfigures from left to right are the reconstructions of the cyan colored points in Fig.~\ref{fig:human_latent}. From some center position (0, 0), the starting point in the latent space is chosen with a radius of 1.5 in the vertical up direction. The direction of movement in the latent space is counterclockwise with a step length of 40 degree. One circle in the latent space corresponds to two steps of a walking movement in the observation space (until the body reaches the same pose as in the beginning).}
	\label{fig:human_1example1}
\end{figure}

The CMU Graphics Lab Motion Capture Database consists of a large number of human motion recordings, which were recorded by using a Vicon motion capture system. Human subjects wear 41 markers while walking, dancing, etc. The data consists of 62-dimensional feature vectors, rendered using Vicon Bodybuilder. We pre-process the 62-dimensional data to 50-dimensional vectors as described in \citep{ChenBUS_2015}. To evaluate the metric on this dataset, we used the walking movements (viz.\ trial 1 to 16) of subject 35, since it is very stable and widely used for algorithm evaluation, e.g. \citep{hm_Schoelkopf2007, hm_Bitzer2008, vaedmp2016}. The total of 6616 frames in the dataset were augmented with Gaussian noise with a standard deviation of 0.03, resulting in four times the size of the original dataset. The noises smooth the latent space which is observed through the magnification factor and the interpolation reconstructions. The architecture of the IWAE ca be found in Table~\ref{table:human}.

Fig.~\ref{fig:human_observation}  and \ref{fig:human_latent} show the geodesic and the Euclidean interpolation\footnote{https://github.com/lawrennd/mocap is used to visualize the movement in the observation space.}. The geodesic follows the path along the data manifold and generates a natural and smooth walking movement. In contrast, the Euclidean interpolation traverses two high $M\!F$ areas which cause large jumps of the movement, while the body poses hardly change in other areas.

\section{Conclusion and future work}
\label{sec:conclusion}


The distance between points in the latent space in general does not reflect the true similarity of corresponding points in the observation space. We gave insight into these issues using techniques from Riemannian geometry, applied to probabilistic latent variable models using neural sampling.

In our approach, the Riemannian distance metric has been successfully applied as an alternative metric that takes into account the underlying manifold. In order to produce shorter distances along the manifold compared to the Euclidean distance, we applied SVD to the metric tensor. 
As a secondary effect, the metric can be used for smoother interpolations in the latent space.

For two-dimensional latent spaces, the $M\!F$ serves as a powerful tool to visualize the magnitude of the generative model's distortion of infinitesimal areas in the latent space. 

%
%

%

%
Future work includes facilitating the use of this distance metric and applying it to models with dynamics such as 
\citep{vaedmp2016} and \citep{DVBF2016}.
%
%
%
%



\bibliographystyle{abbrvnat}
\bibliography{sections/geodesic.bib}

\appendix
 
\appendix

\section{Details of the training procedure}
\label{appendix:training_procedure}

To avoid local minima with narrow spikes of velocity but low overall length, we validate the result during training based on the maximum velocity of Eq.~(\ref{eq:velocity}) and the path length $L+\lambda_{\phi} \max_t \phi(t)$, where $\lambda_{\phi}$ is a hyperparameter.

We found that training with batch gradient descent and the loss defined in Eq.~(\ref{eq:length_approx}) is prone to local minima. 
Therefore, we pre-train the neural network $g_{\nn}$ on $n$ random parametric curves.
As random curves we chose B\'ezier curves \citep{de1986shape} of which the control points are obtained as follows: 
We take $\z_0 \tilde{k}/\tilde{K}+\z_1(\tilde{K}-\tilde{k})/\tilde{K}$, $\tilde{k}=1, 2, \dots, \tilde{K}-1$ as the centers of a uniform distribution, with its support orthogonal to the straight line between $\z_0$ and $\z_1$ and the range $(\z_1-\z_0)/2$.
For each of those random uniforms, we sample once, to obtain a set of $\tilde{K}-1$ random points $\z_{\tilde{k}}$.
Together with $\z_0$ and $\z_1$, these define the control points of the B\'ezier curve.
For each of the $n$ random curves, we fit a separate $g_{\nn}(t)$ to the points of the curve and select the model $g_{\nn}$ with the lowest validation value as the pre-trained model.
Afterwards, we proceed with the optimization of the loss Eq.~\eqref{eq:length_approx}.

\section{Gradients of piecewise linear activation functions}
\label{appendix:gradients}

Note that calculating ${\partial L(g_{\nn}(t))} / {\partial \nn}$ involves calculating the gradients of the Jacobian ${\partial \x} / {\partial \z}$ as well.
Therefore optimization with gradient-based methods is not possible when the generative model uses piecewise linear units.
This can be illustrated with an example of a neural network with one hidden layer:
\begin{align}
\frac{\partial \J}{\partial \z} &=
\frac{\partial \J}{\partial \h} \frac{\partial \h}{\partial \z}
= \frac{\partial}{\partial \h} \Big( \frac{\partial \x}{\partial \h}\frac{\partial \h}{\partial \z} \Big) \frac{\partial \h}{\partial \z} \nonumber \\
&= \frac{\partial^{2} \x}{\partial \h^{2}}\frac{\partial \h}{\partial \z}\frac{\partial \h}{\partial \z} +
\frac{\partial \x}{\partial \h}\frac{\partial^{2} \h}{\partial \z \partial \h}\frac{\partial \h}{\partial \z}.
\label{eq:bpJ}
\end{align}
Both terms in Eq.~(\ref{eq:bpJ}) contain a term that involves twice differentiating a layer with an activation function. In the case of piecewise linear units, the derivative is a constant and hence the second differentiation yields zero. 

\section{Gradients of sigmoid, tanh and softplus activation functions}
\label{appendix:nonlinear_gradient}

We can easily get the Jacobian using sigmoid, tanh and softplus activation functions.
Take one layer with the sigmoid or tanh activation function as an example, the Jacobian is written as
\begin{align}
\J_{i} = \frac{\partial x_i}{\partial \z} = x_i(1-x_i)\w_{i},
\end{align}
where $\w_{i}$ is the weights.

With a softplus activation function, the Jacobian is
\begin{align}
\J_{i} = \frac{\w_{i}}{1+e^{-\z \w_{i} }}.
\end{align}

Consequently, the derivative of Jacobian is straightforward.

\section{Experiment setups}

We used Adam optimizer \citep{corrKingmaB14} for all experiments. FC in the tables refers to fully-connected layers. In Table~\ref{table:mnist}, for the generative model-architecture we used an MLP and residual connections \citep{he2016deep}---additionally to the input and output layer. $K$ is the number of importance-weighted samples in Eq.~(\ref{eq:K}).




\begin{table*}[h]
\caption{The setup for geodesic neural networks} \label{table:geodesic}
\begin{center}
\begin{tabular}{ll}
{\bf architecture } &  {\bf hyperparameters} \\
\hline \\
Input $\in  \mathbb{R}^{N_t}$  & learning rate =  $10^{-2}$ \\
2 tanh FC $\times$ 150 units  & 500 sample points  \\
Output $\in  \mathbb{R}^{N_z}$      &  \\
\end{tabular}
\end{center}
\end{table*}


\begin{table*}[h]
\caption{The setup for the pendulum dataset} \label{table:pendulum}
\begin{center}
\begin{tabular}{lll}
{\bf recognition model}  &{\bf generative model} &{\bf hyperparameters} \\
\hline \\
Input $\in  \mathbb{R}^{256}$  &  Input $\in  \mathbb{R}^{2} $  &  learning rate = $10^{-4}$ \\
2 tanh FC $\times$ 512 units  & 2 tanh FC $\times$ 512 units & $K$ = 50\\
linear FC output layer for means    &  softplus FC output layer for means & batch size = 20\\
softplus FC output layer for variances  & global variable for variances &
\end{tabular}
\end{center}
\end{table*}


\begin{table*}[h]
\caption{The setup for the MNIST dataset} \label{table:mnist}
\begin{center}
\begin{tabular}{lll}
{\bf recognition model}  &{\bf generative model} &{\bf hyperparameters} \\
\hline \\
Input $\in  \mathbb{R}^{784}$  &  Input $\in  \mathbb{R}^{2} $  &  learning rate = $10^{-4}$ \\
2 tanh FC $\times$ 512 units  & 7 residual $\times$ 128 units. & $K = 50$\\
linear FC output layer for means    &  softplus FC output layer for means & batch size = 20\\
softplus FC output layer for variances  & global variable for variances &
\end{tabular}
\end{center}
\end{table*}



\begin{table*}[h]
\caption{The setup for the robot arm simulaiton dataset} \label{table:robot}
\begin{center}
\begin{tabular}{lll}
{\bf recognition model}  &{\bf generative model} &{\bf hyperparameters} \\
\hline \\
Input $\in  \mathbb{R}^{6}$  &  Input $\in  \mathbb{R}^{2} $  & learning rate = $10^{-3}$ \\
2 tanh FC $\times$ 512 units  & 2 tanh FC $\times$ 512 units & $K = 15$\\
linear FC output layer for means    &   softplus FC output layer for means& batch size = 150\\
softplus FC output layer for variances  & global variable for variances &
\end{tabular}
\end{center}
\end{table*}



\begin{table*}[h]
\caption{The setup for the human motion dataset} \label{table:human}
\begin{center}
\begin{tabular}{lll}
{\bf recognition model}  &{\bf generative model} &{\bf hyperparameters} \\
\hline \\
Input $\in  \mathbb{R}^{50}$  &  Input $\in  \mathbb{R}^{2} $  & learning rate = $10^{-3}$ \\
3 tanh FC $\times$ 512 units  & 3 tanh FC $\times$ 512 units & $K = 15$\\
linear FC output layer for means    &   softplus FC output layer for means & batch size = 150\\
softplus FC output layer for variances  & global variable for variances &
\end{tabular}
\end{center}
\end{table*}



\end{document}